\pdfoutput=1
\documentclass[11pt]{article}

\usepackage[preprint]{acl}

\usepackage[T1]{fontenc}
\usepackage[utf8]{inputenc}
\usepackage{microtype}
\usepackage{booktabs}
\usepackage{multirow}
\usepackage{amsmath}
\usepackage{amssymb}
\usepackage{latexsym}
\usepackage{mathptmx}
\usepackage[scaled=0.92]{helvet}  
\usepackage{inconsolata}
\usepackage{bm}      
\usepackage{graphicx}
\usepackage{subcaption}
\usepackage{xcolor}
\usepackage{xspace}
\usepackage{url}
\usepackage{cleveref}
\usepackage{placeins}  
\usepackage{float}     
\usepackage{enumitem}  

\graphicspath{{figures/}}

\newcommand{\method}{IntentKV\xspace}
\newcommand{\phaseone}{Phase-1\xspace}
\newcommand{\learn}{Phase-2\xspace}

\title{IntentKV: Cross-Turn Intent-Aware KV Cache Pruning\\
       for Agent Inference}

\author{%
  Junjie Li \quad Jiong Lou \quad Jie Li \\[2pt]
  Shanghai Jiao Tong University
}

\begin{document}
\maketitle

\begin{abstract}
Multi-turn LLM agents fan short queries into long trajectories of
tool calls, search results, and intermediate reasoning. Both KV
memory and KV read bandwidth grow by orders of magnitude across a
single trajectory, making the key--value (KV) cache, not parameter
compute, the dominant serving bottleneck for long-horizon agents.
We introduce \method, learned KV pruning that keeps the base LLM
frozen. \method maintains a session-level QueryMemory of cross-turn
intent, scores live history tokens with a memory-attention rule,
and adds a zero-initialized residual head with cross-attention over
current-query K-vectors. To stay composable with prefix caches,
eviction is a slot-map redirection: dropped positions route to a
sentinel \emph{dead slot} while surviving K/V rows, RoPE phases,
and slot identities stay in place.
\method matches the no-pruning full-cache baseline with almost no
accuracy drop under tight KV budgets: at an 8k KV budget, mean peak
request tokens drop 23.9\% on Qwen3-8B and 30.7\% on Qwen2.5-14B.
On the 100 longest BCP queries that all methods complete on
Qwen2.5-14B, \method-8k further cuts worst-case peak request tokens
from 92.3k to 20.5k ($\downarrow$ 77.8\%) and worst-case raw KV
reads from 411M to 31M ($\downarrow$ 92.6\%).
\end{abstract}

\section{Introduction}
\label{sec:intro}

\begin{figure*}[t]
\centering
\includegraphics[width=\linewidth]{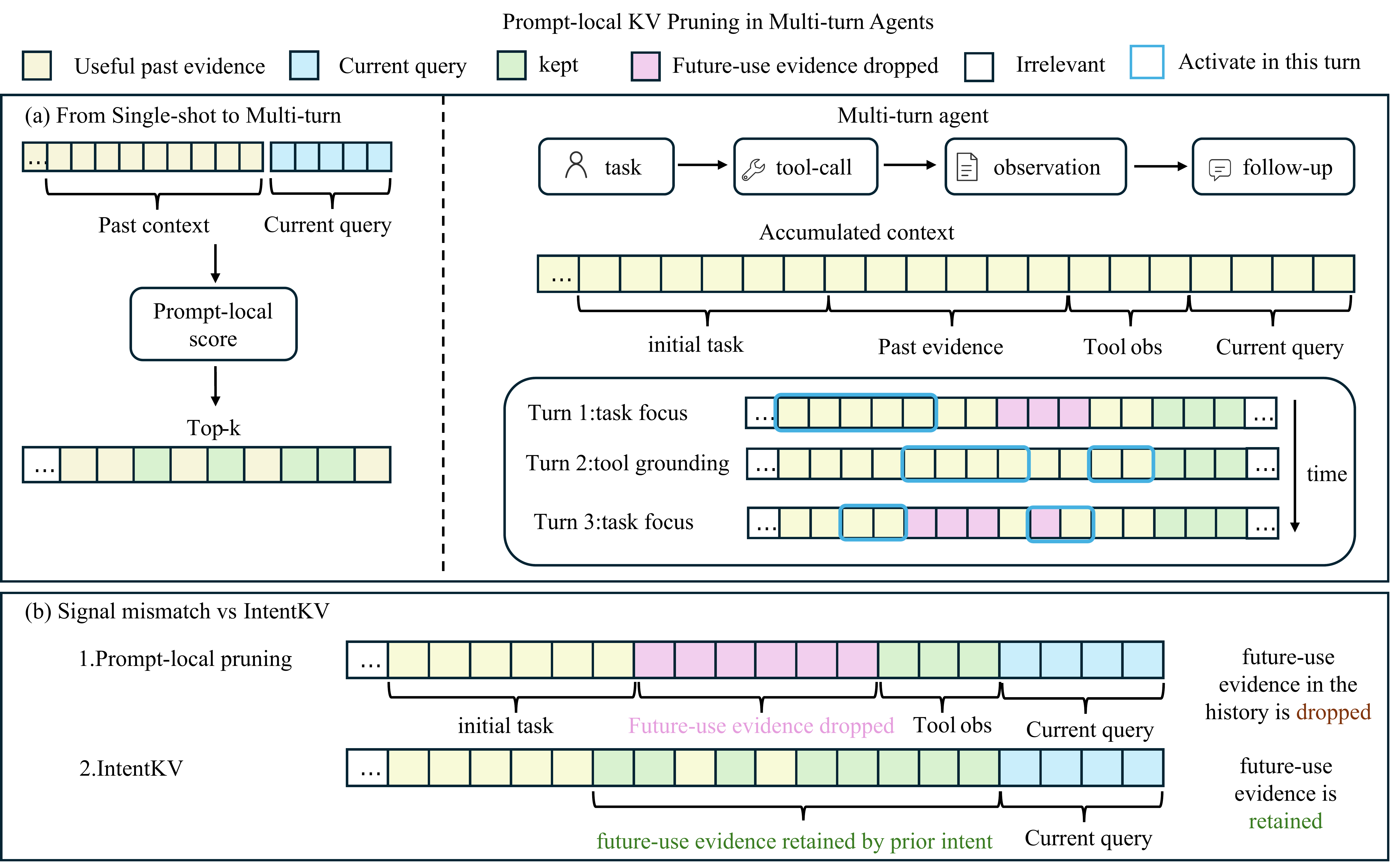}
\caption{\textbf{Prompt-local KV pruning becomes stale in multi-turn
agents.} (a) Single-shot compressors score history against one
query; agent sessions accumulate new intents across turns.
(b) Signal mismatch between prompt-local scoring and \method's
session-aware retention.}
\label{fig:prompt-local-pruning-failure}
\end{figure*}

Modern LLM agents increasingly execute multi-turn browsing,
deep-research, and tool-use workflows~\citep{yao2022react}. In these settings, a short user
request often expands into a long trajectory of retrieved documents,
tool outputs, intermediate decisions, and generated arguments. This
creates an asymmetric serving profile: although the original user
query may contain only a few dozen tokens, the agent state can grow
to tens of thousands of prompt tokens, and each generated token must
attend to the KV tensors of all previous tokens at every layer. As a
result, KV memory capacity and KV read bandwidth, rather than
parameter computation alone, become the dominant bottlenecks in
serving long-running agent sessions
\citep{li2024survey,zheng2024sglang}.

A natural response is to apply existing KV-cache pruning methods, but
most were designed for single-prompt compression rather than
multi-turn agent serving. StreamingLLM~\citep{xiao2024efficient} keeps
attention sinks and a recent window; SnapKV~\citep{li2024snapkv} estimates
token importance from prompt-local tail attention; H2O~\citep{zhang2023h2o}
retains heavy hitters. These methods decide which past tokens to keep from
signals available within a single prompt, and their standard use
typically materializes the kept KV rows as a compacted request-local
cache.

When these methods are applied to multi-turn agent serving, two
assumptions from single-prompt compression become problematic. First,
token importance changes across turns: later queries, retrieved
evidence, and intermediate decisions can make early history useful
again, so a ranking computed from one prompt grows stale and may
discard future-use evidence. Second, the pruned state must remain reusable: agent
sessions repeatedly submit histories with long shared prefixes, which
systems such as SGLang~\citep{zheng2024sglang} and vLLM~\citep{kwon2023efficient} exploit
through radix or prefix caches. Relocating surviving KV rows may
serve the current request, but it changes the cache identity that
later turns would otherwise reuse.
\Cref{fig:prompt-local-pruning-failure} illustrates both failure
modes: prompt-local scoring can drop future-use evidence, and
relocation-based pruning loses the prefix identity needed across
turns.

These two constraints suggest separating what pruning keeps from how
the pruned state is represented. The retention policy should track
information needs as they evolve across the session, rather than
scoring history only against a single prompt. The layout policy should
evict tokens without relocating surviving K/V rows, preserving their
logical positions and cache identities for later turns. We instantiate
this separation in \method: a session-aware heuristic provides the
primary retention signal, a small learned residual handles cases the
rule misses, and a layout-preserving eviction recipe keeps pruning
compatible with prefix-cache reuse.

On the retention side, \method maintains a QueryMemory that
aggregates query-like signals across the session and scores live
history tokens against this evolving state. A small zero-initialized
residual head, trained from frozen-model features, learns a correction
for cases the rule scorer misses while recovering the rule score at
initialization.

On the layout side, \method evicts without compaction. Dropped
positions are redirected to a sentinel \emph{dead slot}, while
surviving K/V rows, RoPE phases, and logical slot identities remain
unchanged. Agent sessions can therefore combine KV pruning with
cross-turn prefix reuse, rather than choosing one or the other.

\paragraph{Contributions.}
\begin{itemize}[itemsep=1pt,parsep=0pt,topsep=2pt]
    \item We formulate agent KV pruning as \emph{multi-query
    retention} and instantiate it with QueryMemory, a per-session
    state aggregating queries, tool calls, search intents, and action
    spans across turns.
    \item A learned-residual pruner keeps the session-aware
    heuristic as the primary signal and adds a frozen-feature
    correction grounded in future actions, without updating the base
    LLM.
    \item An eviction recipe for paged KV caches (SGLang/vLLM) that
    redirects dropped positions to a sentinel slot rather than
    relocating survivors, so pruning stays composable with
    radix-prefix reuse without changes to existing attention kernels:
    $20.7\%$ prefix-hit rate at an $8$k budget where compaction
    baselines fall to $0$--$3\%$.
    \item On BCP at $C{=}8$k, \method-\learn matches Full-cache True
    Acc within $0.96$ points on Qwen3-8B and surpasses the strongest
    heuristic by $10.36$ points on Qwen2.5-14B; on the $100$ longest
    queries it cuts worst-case peak request tokens by
    $77.8\%$--$81.7\%$ and worst-case raw KV reads by
    $36.9\%$--$92.6\%$ (\Cref{tab:worst-case-kv}).
\end{itemize}

\section{Related Work}
\label{sec:related}

\paragraph{From prompt-local scores to cross-turn intent.}
Most KV eviction methods score tokens against a single user
prompt~\citep{xiao2024efficient,liu2023scissorhands,zhang2023h2o,li2024snapkv,zhou2024dynamickv,ahn2026lookaheadkv},
with orthogonal refinements for per-layer budgets, query-aware
decode-time sparsity, and future-query
pre-scoring~\citep{ge2024model,cai2024pyramidkv,tang2024quest,devoto2025expected}. All assume a single
request available at prefill. Multi-turn agents break this: later
turns surface fresh query signals in tool outputs and intermediate
plans that a one-shot scorer cannot anticipate. \method instead
formulates pruning as multi-query retention, scoring history against
a session-level QueryMemory that absorbs these signals as they
arrive. KV quantization~\citep{liu2024kivi,kang2024gear} reduces per-token precision
rather than token count and is orthogonal to \method.

\paragraph{Semantic pruning without changing the base model.}
A second line learns token-importance policies on frozen-model
features~\citep{ahn2026lookaheadkv,zhou2024dynamickv}, though pure learned scores can
be brittle under weak or shifted query signal.
TRIM-KV~\citep{bui2025cache} is the closest peer: it trains a small
predictor and biases retention via a \emph{positional time-decay}
prior, well-suited to chain-of-thought trajectories where relevant
context is temporally local. Agent loops invert this regime---useful
tokens are often the oldest (initial request, early retrieved page,
first tool result) while youngest decode tokens are scaffolding---so
a time-decay prior is tuned to a different workload. \method trains a
comparable head but conditions it on QueryMemory and supervises with
which positions are re-attended in later turns, yielding a
\emph{query-decay} policy anchored to evolving intent.
SideQuest~\citep{kariyappa2026sidequest} and Activation Beacon~\citep{zhang2025long}
modify the base model (deletion commands and learned summary tokens
respectively) and so are complementary rather than substitutable;
\method keeps the base LLM frozen and adds only a zero-initialized
residual head.

\paragraph{Composable serving- and orchestration-layer systems.}
Efficient agent serving reuses KV state via paged allocation and
prefix/radix caches when consecutive requests share a
prefix~\citep{kwon2023efficient,zheng2024sglang}, and Continuum~\citep{li2025continuum} extends
this to multi-turn agents through KV time-to-live scheduling.
Conventional compaction shifts the physical slots and RoPE phases of
surviving tokens and invalidates the prefix identity such systems
would reuse; \method keeps surviving K/V rows in place and redirects
dropped logical positions to a sentinel slot, so request-local
compression composes with radix-prefix reuse. At a higher layer,
MemGPT~\citep{packer2023memgpt} and A-MEM~\citep{xu2026mem} manage agent state via
OS-style swapping and long-term graph memory, deciding what to
remember at the orchestration layer; \method operates inside the
cache with the base LLM and agent loop unchanged and so composes with
such memories rather than competing.

\section{Method}
\label{sec:method}

\begin{figure*}[!t]
\centering
\includegraphics[trim=0 3 207.5 31.5,clip,width=\linewidth]{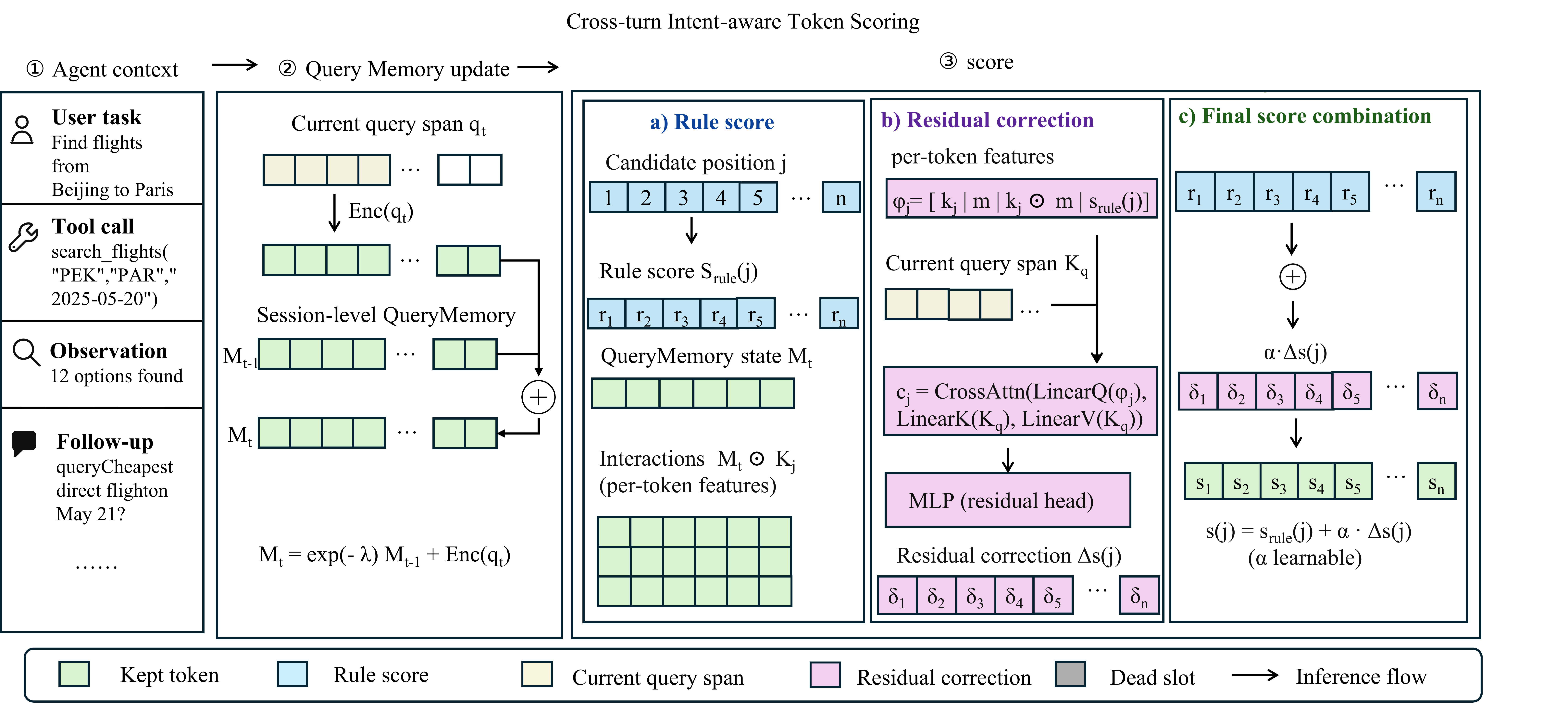}
\caption{\textbf{\method method overview.} (a) Rule score against
QueryMemory; (b) zero-init residual correction via cross-attention
over current-query K-vectors; (c) final score combination feeding
top-$k$ retention. Eviction redirects dropped positions to a
sentinel \emph{dead slot} on the existing paged KV cache, preserving
radix prefix identity.}
\label{fig:method-overview}
\end{figure*}

\method (\Cref{fig:method-overview}) intervenes once per request
after prefill, when the sequence length $N$ exceeds a retention
budget $C$. It produces a kept set
$\mathcal{K}\!\subseteq\![0,N)$ that always retains the current
actionable query span and the protected system prefix, holds every
other surviving token at its original logical position and physical
KV slot, and lets the cross-request radix tree~\citep{zheng2024sglang} match
prefixes despite eviction. The design factors into two layers:
\emph{retention} ranks history under cross-turn intent, and
\emph{layout} keeps the surviving prefix reusable.

\subsection{Problem Formulation}
\label{sec:method-problem}

We frame KV pruning as a budgeted top-$k$ selection over post-prefill
tokens, with a small protected set fixed in advance. Let
$\mathbf{S}_r\!\in\!\mathbb{Z}^N$ denote the slot map of request
$r$, $\mathbf{K},\mathbf{V}\!\in\!\mathbb{R}^{L\times H_{kv}\times
N\times D}$ the post-RoPE~\citep{su2024roformer} keys and values, and $[q_s,q_e)$ the
actionable query span (the latest user, tool, or function message
that can affect the next decision, resolved after template
rendering). The protected prefix $\pi$ covers the chat-template
system span. With the forced set
$\mathcal{F}\!=\![0,\pi)\cup[q_s,q_e)$, residual budget
$C^{\star}\!=\!\max(0,C-|\mathcal{F}|)$, and candidate history
$\mathcal{H}\!=\![0,N)\setminus\mathcal{F}$, compression selects
\begin{equation}
\mathcal{K}^{\star}=\mathcal{F}\,\cup\,
\operatorname*{arg\,top}_{\substack{\mathcal{K}\subseteq\mathcal{H}\\ |\mathcal{K}|=C^{\star}}}
\sum_{j\in\mathcal{K}} s_j,
\label{eq:topk}
\end{equation}
where $s_j$ is produced by one of three scorer modes:
\texttt{query} uses only the current query encoding,
\texttt{memory} accumulates queries across the session via a
recurrence, and \texttt{learnable} adds a residual head on top of
\texttt{memory}. Thus $C$ controls compressible history rather
than serving as a hard cap on prompt-plus-generation tokens;
implementation details are in \Cref{appendix:algos}.

\subsection{Cross-Turn Query Memory}
\label{sec:method-memory}

\begin{figure}[t]
\centering
\includegraphics[width=\linewidth]{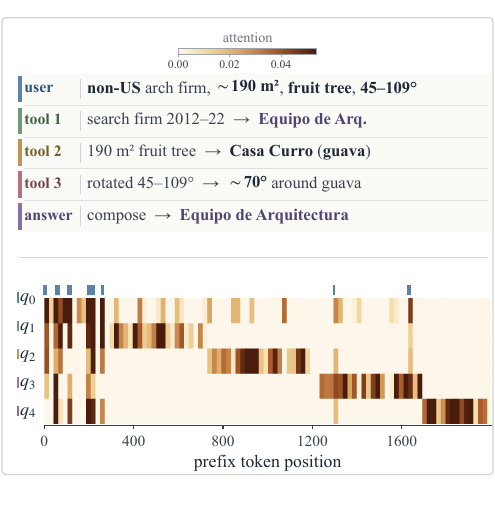}
\caption{\textbf{Cross-turn attention drifts away from the initial
prompt.} On a BCP rollout (qid~$861$, Qwen3-8B; turn-aware synthetic
preview), each query $q_0\!\to\!q_4$ activates a different region of
the prefix (positions $0$--$2000$).}
\label{fig:motivation-drift}
\end{figure}

Within one agent session, attention from later queries no longer
concentrates on the regions that the first user prompt singled out,
so a scorer trained on $q_0$ alone is mis-calibrated for $q_{1{:}t}$
(\Cref{fig:motivation-drift}). Per session, \method maintains a
query memory
$\mathbf{M}_t\!\in\!\mathbb{R}^{L\times H_q\times D}$ with the shape
of one post-RoPE query row, updated by a geometric recurrence:
\begin{equation}
\mathbf{M}_t = e^{-\lambda}\,\mathbf{M}_{t-1} +
\mathrm{Enc}(\mathbf{q}_t),
\label{eq:memory-update}
\end{equation}

where $\mathrm{Enc}(\mathbf{q}_t)$ averages the post-RoPE Q rows
over the actionable query span; the decay rate $\lambda$ is fixed
across the session (value set in \Cref{appendix:hyperparams}).
After each update, $\mathbf{M}_t$ is projected to unit norm along
the head dimension; without this, its norm grows roughly linearly
with turn count and the softmax below collapses onto a single
coordinate. With $\mathbf{K}\!\in\!\mathbb{R}^{L\times H_{kv}\times
N\times D}$ from the current request and KV heads expanded to
$H_q$ under GQA, candidate position $j\!\in\!\mathcal{H}$ receives
the rule score
\begin{equation}
\begin{aligned}
\mathrm{rule}_j
&= \sum_{l,h} a_{l,h,j},\\
a_{l,h,j}
&=
\frac{\exp(\mathbf{M}_t[l,h]\!\cdot\!\mathbf{K}[l,h,j]/\sqrt{D})}
     {\sum_{i\in\mathcal{H}}\exp(\mathbf{M}_t[l,h]\!\cdot\!\mathbf{K}[l,h,i]/\sqrt{D})}.
\end{aligned}
\label{eq:rule-score}
\end{equation}
\Cref{eq:rule-score} replaces SnapKV's window of $W$ tail Q rows
with a single accumulated query state, reducing per-event scoring
from $\mathcal{O}(W L H_q N)$ to $\mathcal{O}(L H_q N)$ while
incorporating evidence from prior turns.

\subsection{Residual Learnable Pruner}
\label{sec:method-learn}

We refer to the rule scorer of \Cref{eq:rule-score} as the
\phaseone configuration and to the rule-plus-residual scorer
defined below as the \learn configuration.
\method optionally augments the rule score with a learned residual
that produces the $s_j$ consumed by \Cref{eq:topk},
\begin{equation}
s_j = \mathrm{rule}_j +
\alpha\cdot\mathrm{MLP}\!\big(\bm{\phi}_j\oplus\mathbf{c}_j\big),
\label{eq:residual}
\end{equation}
where $\alpha\!\in\![-5,5]$ is a learned scalar clipped at
inference, $\bm{\phi}_j\!\in\!\mathbb{R}^{3D+1}$ is the per-token
feature vector, and $\mathbf{c}_j\!\in\!\mathbb{R}^{d_c}$ is a
cross-attention readout over the current query's K-vectors.

\paragraph{Per-token feature vector.}
Let $\bar{\mathbf{k}}_j$ be the mean of $\mathbf{K}[l,h,j]$ over
GQA-expanded heads and layers, and $\bar{\mathbf{m}}$ the analogous
mean over $\mathbf{M}_t$. Then
\begin{equation}
\bm{\phi}_j=
\big[\,\bar{\mathbf{k}}_j\,\Vert\,\bar{\mathbf{m}}\,\Vert\,
\bar{\mathbf{k}}_j\!\odot\!\bar{\mathbf{m}}\,\Vert\,\mathrm{rule}_j\,\big].
\end{equation}
Mean aggregation keeps every column of $\bm{\phi}_j$ at
$\mathcal{O}(1)$ regardless of depth or head count, so one residual
topology covers backbones with the same $D$. The element-wise
interaction $\bar{\mathbf{k}}_j\!\odot\!\bar{\mathbf{m}}$ supplies
an explicit token--memory alignment channel that a linear
concatenation cannot represent.

\paragraph{Cross-attention over query K-vectors.}
Per-token features collapse every layer and head, losing which
current query token a candidate matches. The residual therefore
reads the query span back through a multi-head cross-attention. Let
$\bar{\mathbf{K}}^{q}\!\in\!\mathbb{R}^{n_q\times D}$ aggregate
K-vectors at the live query positions (positions evicted in a prior
round are excluded, so the attention consumes only live evidence). With $H_c\!=\!4$ heads and width
$d_c\!=\!128$,
\begin{equation}
\mathbf{c}_j=\mathrm{CrossAttn}\big(W_Q\bm{\phi}_j,\;
W_K\bar{\mathbf{K}}^{q},\;W_V\bar{\mathbf{K}}^{q}\big).
\label{eq:xattn}
\end{equation}
Candidate features form the queries; past query K-vectors form
keys and values.

\paragraph{Initialisation and lower bound.}
The MLP body is a two-layer GELU network of hidden width
$h\!=\!256$. Its output projection is zero-initialised in both
weight and bias, while $\alpha\!=\!1$ at start. Under
$(\alpha\!=\!1,\,\mathrm{fc}_2\!=\!0)$ the pruner exactly recovers
\Cref{eq:rule-score} on day zero, gradients still flow through
$\mathrm{fc}_2$, and the rule scorer remains a strict floor under
the learned head; pairing $\alpha\!=\!0$ with $\mathrm{fc}_2\!=\!0$
would freeze the gradient path at a $(0,0)$ saddle. For
$D\!=\!128$, the residual head has $214{,}274$ trainable
parameters.

\paragraph{Ablation knobs.}
The full configuration \emph{Full \learn} activates all three
additions in \Cref{eq:residual}. \Cref{sec:experiments-ablation}
ablates them independently: \emph{$-$~memory} replaces
$\mathbf{M}_t$ with $\mathrm{Enc}(\mathbf{q}_t)$,
\emph{$-$~x-attn} forces $\mathbf{c}_j\!=\!\mathbf{0}$ in
\Cref{eq:xattn}, and \emph{$-$~residual} clamps $\alpha\!=\!0$,
exactly recovering the rule scorer.

\subsection{Slot-Map Eviction}
\label{sec:method-layout}

Serving stacks already decouple logical token positions from
physical KV slots through a paged allocator and a per-request slot
map~\citep{kwon2023efficient,zheng2024sglang}. Standard compactors copy kept rows into a
contiguous range and re-rotate keys, which changes both slot
identity and RoPE phase and breaks the radix-tree match against
later turns. \method instead leaves every kept K/V row at its
original slot and modifies only $\mathbf{S}_r$:
\begin{equation}
\mathbf{S}_r[j]\leftarrow s^{\dagger},\quad
\forall j\in[0,N)\setminus\mathcal{K}^{\star},
\label{eq:redirect}
\end{equation}
with the token-id stream $\mathbf{fill\_ids}_r$ and sequence length
$N$ untouched. The radix tree therefore continues to match this
request against later turns sharing its prefix, exactly as if no
pruning had happened.

\paragraph{Sentinel as in-place mask.}
The redirect target $s^{\dagger}$ in \Cref{eq:redirect} is a single
sentinel slot allocated at compressor startup and held in a
never-free set, with
$\mathbf{K}[s^{\dagger}]\!=\!-10^{4}\mathbf{1}_D$ on every layer and
$\mathbf{V}[s^{\dagger}]\!=\!\mathbf{0}_D$. The pre-softmax logit at
the sentinel is thus $-10^{4}\sqrt{D}$, driving its softmax weight
to $\exp(-10^{4}\sqrt{D})$, far below the smallest representable
bf16/fp16 value; the zero value then nulls any residual
contribution. The sentinel is numerically equivalent to a hard
visibility mask, but it lives at the slot-map level rather than
inside the attention kernel, so existing flashinfer and FA3 paths
run unmodified.

\paragraph{Alias-aware deallocation.}
A dropped physical slot $u\!=\!\mathbf{S}_r[j]$ is returned to the
allocator only if (a) position $j$ lies outside the radix-protected
prefix, (b) $u\!\neq\!s^{\dagger}$, and (c) $u\notin
\{\mathbf{S}_r[i]:i\!\in\!\mathcal{K}^{\star}\}$. Condition (c)
catches two alias patterns: positions already redirected to
$s^{\dagger}$ in a prior compression round, and slots shared with
sibling requests on the same radix branch. Without this check,
freeing $u$ would corrupt KV reads from those siblings, which is
the practical obstacle that forces prior compaction-based pruners
to disable prefix-cache reuse. The slot-map indirection is
borrowed from PagedAttention~\citep{kwon2023efficient} and the sentinel-as-mask
construction is the standard attention-masking idiom; what we add
is the integration itself: combining them so per-request pruning
becomes composable with radix-prefix reuse without modifying the
attention kernel.

\subsection{Training}
\label{sec:method-train}

\paragraph{Frozen-model feature extraction.}
Training reads post-RoPE Q and K from a frozen copy of the deployed
model. We register a RoPE hook that captures the post-rotation
tensors exactly where the attention backend consumes them, so the
offline feature distribution matches inference. Long sequences are
processed in $4{,}096$-token chunks with KV caching enabled to
bound attention-matrix memory; chunk boundaries respect the
actionable query span so each example sees its full encoding.

\paragraph{Future-action grounding.}
For each $(\texttt{session},\,\texttt{turn})$ row, the labeller
scans the next five tool calls in the same session and marks any
history token whose content matches a literal substring of a future
tool argument as positive. Tokens inside the current query span and
special-template tokens carry label $-100$ and are excluded from
the loss; rows with no matched evidence are dropped rather than
back-filled with a recency proxy, which improves label precision at
the cost of some textually unrecoverable turns.

\paragraph{Loss.}
The total objective sums four terms,
\begin{equation}
\begin{aligned}
\mathcal{L}
&=\mathcal{L}_{\text{BCE}}(s,y)+\rho\,\mathcal{L}_{\text{rank}}\\
&\quad+\gamma\,\overline{\sigma(s)}+\eta\,(\Vert\mathbf{M}_t\Vert_2-1)^{2},
\end{aligned}
\label{eq:loss}
\end{equation}
with defaults $\rho\!=\!0.05$, $\gamma\!=\!0.01$, $\eta\!=\!0.001$.
$\mathcal{L}_{\text{BCE}}$ uses an adaptive positive weight
$(1-\tilde p)/\tilde p$ with
$\tilde p\!=\!\mathrm{clip}(p,0.02,0.5)$ to absorb the long-tailed
positive rate; $\mathcal{L}_{\text{rank}}$ is a pairwise
$\operatorname{softplus}$ surrogate that compares each positive
against the top-$64$ hard negatives; the third term discourages an
unbounded keep probability; the last term softly enforces the
unit-norm projection on $\mathbf{M}_t$ used at inference.

\paragraph{Optimisation.}
Sessions are replayed in increasing \texttt{turn\_index} so
$\mathbf{M}_t$ accumulates as at inference and are shuffled between
epochs. One residual head is trained per backbone; the topology is
identical and the differences are captured by separately trained
weights. See \Cref{sec:experiments-setup} for hyperparameters.

\section{Experiments}
\label{sec:experiments}

\begin{figure*}[!t]
\centering
\includegraphics[width=0.95\linewidth]{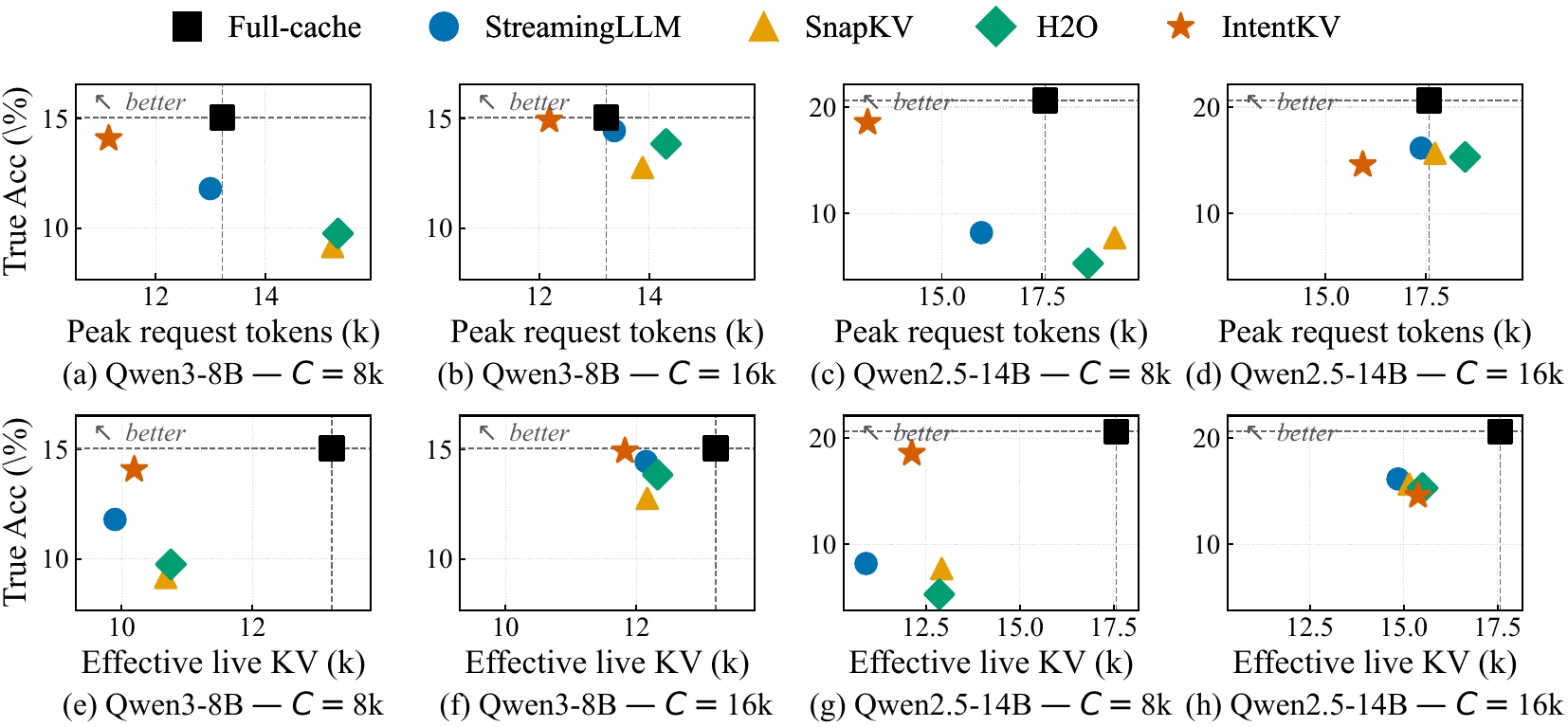}
\caption{\textbf{BCP accuracy under two distinct cost metrics.}
Each column fixes one (model, budget) configuration. The top row
plots True~Acc against the \emph{KV memory footprint} (PT, k
tokens). The bottom row plots True~Acc against the
\emph{decode-side attention workload} (\textbf{Eff. Live KV}, k
tokens). \method-\learn is Pareto-dominant on both axes at
$C{=}8$k.}
\label{fig:main-pareto}
\end{figure*}

\subsection{Experimental Setup}
\label{sec:experiments-setup}

\paragraph{Models, serving, and benchmark.}
We evaluate Qwen3-8B-Instruct~\citep{yang2025qwen3} and
Qwen2.5-14B-Instruct~\citep{qwen2025qwen25technicalreport} on a single $80$\,GiB A100,
both Neox-RoPE GQA models. The restriction to Qwen backbones
is empirically motivated rather than methodological: a seven-model
cross-architecture smoke study spanning Llama-3.1-8B,
Mistral-Nemo-2407, Mistral-Small-3.2-24B, Hermes-3-Llama-3.1-8B,
watt-tool-8B, gpt-oss-20b, and GLM-4-9B found that no non-Qwen
open-source backbone in our sample reliably engages in autonomous
multi-step tool calling
on BCP, even after we fixed the corresponding parser and
tokenizer-protocol incompatibilities; the failures sit at the
agentic-behavior layer, not at the KV-compression layer
(\Cref{appendix:cross-architecture}). All inference uses
SGLang~\citep{zheng2024sglang} with radix prefix caching, deterministic
decoding ($T{=}0$), and at most $32$ tool-use turns per query.
The benchmark is BrowseComp-Plus (BCP;~\citealp{chen2025browsecomp}), an
$830$-query deep-research suite over a fixed $\sim$100K-document
corpus; the retriever is frozen via a per-query cache so compression
effects are not confounded by retrieval drift.

\paragraph{Training.}
The \learn pruner is trained on strict-cleaned ToolBench~%
\citep{qin2024toolllm,guo2024stabletoolbench} multi-turn traces with future-action
labels obtained by literal substring matching against the next five
tool-call arguments (\Cref{sec:method-train}). Training uses
reservoir-sampled $3$K--$5$K examples per epoch for two epochs with
AdamW (lr $10^{-3}$, weight decay $0.01$, grad-accum $8$, seed $42$);
the base LLM is frozen in bf16 and only the FP32 residual head is
updated. Trigger size is $512$ tokens and tail window $W{=}32$. The
Qwen3-8B (resp.\ Qwen2.5-14B) run completes in $\sim$$40$
($\sim$$55$) minutes on $4{\times}80$\,GiB A100s with final loss
$0.54$ ($0.51$); training curves are in
\Cref{appendix:training-curves}.

\paragraph{Baselines.}
We compare \method-\learn against a no-pruning \textbf{Full-cache}
ceiling, \textbf{StreamingLLM}~\citep{xiao2024efficient} (sink $4$ +
recent window), \textbf{SnapKV}~\citep{li2024snapkv} (kernel $7$, $W{=}32$,
max-pool), \textbf{H2O}~\citep{zhang2023h2o} (ratio mode with recent floor),
and a public \textbf{TrimKV}~\citep{bui2025cache} checkpoint that uses
positional time-decay rather than query-conditioned retention. All
compressors run at $C\in\{8192,16384\}$ applied to the compressible
history at each pruning event.

\paragraph{Metrics.}
\emph{Accuracy.} Following BCP, a Qwen3-32B LLM
judge~\citep{zheng2023judging} labels each completed trajectory; we report \textbf{Compl} (queries finishing without OOM,
overflow, turn cap, error, or empty output), \textbf{Raw}
($\textsc{Correct}/\textsc{Completed}$), and the headline
\textbf{True}~Acc ($\textsc{Correct}/\textsc{Total}=$
Raw$\times$Compl, with non-completion counted as wrong).
\emph{System cost.} All system-side metrics are lower-is-better. Let
$R_t$ denote the raw (pre-pruning) attended-to set at decode step
$t$ and $K_t\subseteq R_t$ the \emph{effective} subset that survives
pruning and actually participates in attention; $|R_t|$ and $|K_t|$
are the corresponding sizes. We report:
\begin{itemize}[itemsep=1pt,parsep=0pt,topsep=2pt]
  \item \textbf{PT} (peak request tokens, k): per-query maximum
        $\max_t |R_t|$ of the uncompressed request size; a
        memory-footprint upper bound.
  \item \textbf{Eff. Live KV} (k tokens): per-query peak of the
        live (post-pruning) effective KV, $\max_t |K_t|$; the
        attention-workload metric reported on the bottom row of
        \Cref{fig:main-pareto}.
  \item \textbf{Raw KV Reads} ($10^{6}$ tokens): cumulative
        pre-pruning KV reads $\sum_t |R_t|$.
  \item \textbf{Eff. KV Reads} ($10^{6}$ tokens): cumulative
        effective KV reads $\sum_t |K_t|$; proxies attention
        bandwidth.
\end{itemize}
We additionally report wall-clock latency (\textbf{Wall}, s) and the
radix prefix-hit rate (in \%, quoted inline). Cohort statistics in
\Cref{tab:system-efficiency,tab:abl-components,tab:design-axis} are
means across the relevant query set; \Cref{tab:worst-case-kv}
reports cohort maxima $PT_{\max}{=}\max_q \mathrm{PT}(q)$ and
$R_{\max}{=}\max_q \mathrm{Raw\,KV\,Reads}(q)$.

\subsection{Main Results: Accuracy}
\label{sec:experiments-accuracy}

\paragraph{The 8k budget is the stress regime.}
\Cref{tab:main-accuracy} reports the full $830$-query evaluation.
On Qwen3-8B at $C{=}8$k, the four compaction baselines collapse on
completion: StreamingLLM finishes $65.42\%$ ($-23.6$ vs.\
Full-cache); H2O, SnapKV, TrimKV drop further to $45.66$--$51.57\%$.
\method-\learn keeps $84.58\%$ completion and reaches $14.10$
True~Acc, surpassing StreamingLLM by $2.29$, TrimKV by $3.14$, H2O
by $4.34$, SnapKV by $4.94$. The Qwen2.5-14B margin is larger:
\method-\learn at $C{=}8$k recovers $18.55$ True~Acc vs.\ a
$20.60$ Full-cache ceiling, while the best heuristic (StreamingLLM)
reaches only $8.19$—a $10.36$-point gap.

\paragraph{At 16k the gap to the ceiling closes.}
On Qwen3-8B at $C{=}16$k, \method-\learn attains $14.94$ True~Acc,
within $0.12$ of no-pruning and ahead of every heuristic
(StreamingLLM $14.46$, H2O $13.86$, TrimKV $13.25$, SnapKV $12.77$).
On Qwen2.5-14B the margin narrows ($14.58$ vs.\ StreamingLLM
$16.14$): Qwen2.5-14B BCP trajectories average $\sim$$17$k peak
request tokens, so a $16$k budget rarely triggers eviction—
consistent with the same model's $10.36$-point \method advantage at
the tighter $8$k budget where retention decisions actually matter.

\begin{table}[!htbp]
\centering
\footnotesize
\caption{\textbf{\method-\learn lands within $0.96$ True~Acc of
Full-cache on Qwen3-8B and beats the strongest heuristic by
$10.36$ points on Qwen2.5-14B} ($830$-query BCP eval). Best
non-Full-cache cell per column in \textbf{bold}, second-best
\underline{underlined}; Full-cache is shown only at $C{=}16$k as
the uncompressed ceiling.}
\label{tab:main-accuracy}
\setlength{\tabcolsep}{2pt}
\renewcommand{\arraystretch}{1.05}
\begin{tabular*}{\columnwidth}{@{\extracolsep{\fill}} @{}l rrr rrr@{}}
\toprule
 & \multicolumn{3}{c}{$C{=}8$k} & \multicolumn{3}{c}{$C{=}16$k} \\
\cmidrule(lr){2-4}\cmidrule(lr){5-7}
Method & Raw & Compl & True & Raw & Compl & True \\
\midrule
\multicolumn{7}{l}{\emph{Qwen3-8B}} \\
Full-cache    & ---   & ---   & ---   & 16.92 & 89.04 & 15.06 \\
StreamingLLM  & 18.05 & 65.42 & \underline{11.81}
              & 18.10 & 79.88 & \underline{14.46} \\
SnapKV        & 20.06 & 45.66 & 9.16
              & 17.29 & 73.86 & 12.77 \\
H2O           & 21.37 & 45.66 & 9.76
              & 19.36 & 71.57 & 13.86 \\
TrimKV        & 21.25 & 51.57 & 10.96
              & 18.42 & 71.93 & 13.25 \\
\method-\learn & 16.67 & \textbf{84.58} & \textbf{14.10}
               & 17.08 & \textbf{87.47} & \textbf{14.94} \\
\midrule
\multicolumn{7}{l}{\emph{Qwen2.5-14B}} \\
Full-cache    & ---   & ---   & ---   & 21.03 & 97.95 & 20.60 \\
StreamingLLM  & 16.54 & 49.52 & \underline{8.19}
              & 21.61 & 74.70 & \textbf{16.14} \\
SnapKV        & 23.44 & 32.89 & 7.71
              & 21.41 & 73.13 & \underline{15.66} \\
H2O           & 19.30 & 27.47 & 5.30
              & 22.24 & 68.80 & 15.30 \\
\method-\learn & 20.53 & \textbf{90.36} & \textbf{18.55}
               & 20.48 & 71.20 & 14.58 \\
\bottomrule
\end{tabular*}
\end{table}

\subsection{Main Results: Efficiency}
\label{sec:experiments-efficiency}

\paragraph{Efficiency gains come from both tighter retention and
preserved prefix reuse.}
\Cref{tab:system-efficiency} shows \method's Eff. KV Reads sits at
or below every compaction baseline on both backbones: it is within
$8$M of StreamingLLM on Qwen3-8B and roughly $2\times$ lower than
every compaction baseline on Qwen2.5-14B ($21.0$--$24.1$ vs.\
$44.5$--$58.7$M), indicating that QueryMemory selects a tighter
live working set than tail-attention or heavy-hitter heuristics.
On top of this, the hole-preserving layout keeps radix prefix
caches valid: $20.7\%$/$26.0\%$ prefix-hit at $C{=}8$k/$16$k,
while compaction baselines fall to $0$--$3\%$ because relocation
renumbers positions. With both effects active, \method-\learn at
$C{=}16$k on Qwen3-8B issues $32.0$M Raw KV Reads (matching
Full-cache's $32.2$M; $-51.1\%$ vs.\ H2O's $65.5$M), with wall
time $139.2$ s ($-43.1\%$ vs.\ H2O, within $5.9\%$ of no-pruning).
At $C{=}8$k, Raw KV Reads drop to $31.6$M ($-59.1\%$ vs.\
H2O/SnapKV). The prefix-reuse contribution is isolable in
\Cref{appendix:layout-ablation}: re-running SnapKV and H2O on
\method's dead-slot substrate cuts wall-time $44$--$46\%$ and Raw
KV Reads $39$--$47\%$ on both baselines with no change to scoring
or budget, while \method-\learn retains a $1.5\times$ reduction in
Raw KV Reads over SnapKV under the matched substrate at
indistinguishable True~Acc ($\Delta{=}0.35\sigma$ on $830$ queries).

\paragraph{Worst-case KV pressure (vs.\ uncompressed).}
\Cref{tab:worst-case-kv} stress-tests \method against the
\emph{uncompressed} ceiling on the $100$ BCP queries with the
largest Full-cache $PT_{\max}$. \method-$8$k cuts worst-case peak
request tokens by $77.8$--$81.7\%$ and worst-case Raw KV Reads by
$36.9$--$92.6\%$ across the two backbones, while True~Acc stays
within $0.96$--$2.05$ points of the uncompressed ceiling.
\method therefore absorbs the heaviest BCP trajectories at a
$1$--$2$ order-of-magnitude smaller KV footprint without
sacrificing full-cache accuracy.

\begin{table}[!htbp]
\centering
\footnotesize
\caption{\textbf{\method-\learn cuts wall time by up to $43\%$
and Raw KV reads by up to $59\%$ versus compaction baselines on
both backbones} (lower is better). Column definitions in
\Cref{sec:experiments-setup}; best non-ceiling cell per column in
\textbf{bold}.}
\label{tab:system-efficiency}
\setlength{\tabcolsep}{4pt}
\begin{tabular*}{\columnwidth}{@{\extracolsep{\fill}} l rrrr}
\toprule
Method & PT (k) & Wall & Raw Rd & Eff. Rd \\
\midrule
\multicolumn{5}{l}{\emph{Qwen3-8B, $C{=}8$k}} \\
StreamingLLM    & 8.4          & 180.6          & 54.3          & 45.2 \\
SnapKV          & 9.0          & 279.2          & 77.1          & 59.4 \\
H2O             & 9.1          & 281.4          & 77.2          & 59.7 \\
\method-\learn  & \textbf{8.1} & \textbf{140.4} & \textbf{31.6} & \textbf{39.8} \\
\midrule
\multicolumn{5}{l}{\emph{Qwen3-8B, $C{=}16$k}} \\
Full-cache      & 10.6         & 131.4          & 32.2          & 45.1 \\
StreamingLLM    & \textbf{9.9} & 165.3          & 50.0          & 51.5 \\
SnapKV          & \textbf{9.9} & 228.2          & 60.7          & 59.1 \\
H2O             & 10.2         & 244.7          & 65.5          & 63.6 \\
\method-\learn  & \textbf{9.9} & \textbf{139.2} & \textbf{32.0} & \textbf{43.2} \\
\midrule
\multicolumn{5}{l}{\emph{Qwen2.5-14B, $C{=}8$k}} \\
StreamingLLM    & \textbf{10.5} & 275.3          & 68.3          & 44.5 \\
SnapKV          & 12.7          & 574.2          & 101.0         & 56.5 \\
H2O             & 12.6          & 373.1          & 93.6          & 52.4 \\
\method-\learn  & 11.8          & \textbf{104.0} & \textbf{26.7} & \textbf{21.0} \\
\midrule
\multicolumn{5}{l}{\emph{Qwen2.5-14B, $C{=}16$k}} \\
Full-cache      & 17.1          & 203.8          & 25.2          & 27.1 \\
StreamingLLM    & \textbf{14.5} & 292.4          & 64.5          & 51.8 \\
SnapKV          & 14.9          & 447.5          & 70.7          & 52.6 \\
H2O             & 15.2          & 521.3          & 80.9          & 58.7 \\
\method-\learn  & 15.1          & \textbf{105.0} & \textbf{28.8} & \textbf{24.1} \\
\bottomrule
\end{tabular*}
\end{table}

\begin{table}[!htbp]
\centering
\footnotesize
\caption{\textbf{\method-$8$k cuts worst-case peak request tokens
by $78$--$82\%$ and worst-case Raw KV reads by up to $93\%$ against
the uncompressed ceiling.} Cohort: the $100$ BCP queries with the
largest uncompressed $PT_{\max}$ where all three methods complete;
symbols defined in \Cref{sec:experiments-setup}.}
\label{tab:worst-case-kv}
\setlength{\tabcolsep}{3pt}
\begin{tabular*}{\columnwidth}{@{\extracolsep{\fill}} l rrr rrr}
\toprule
 & \multicolumn{3}{c}{Qwen3-8B} & \multicolumn{3}{c}{Qwen2.5-14B} \\
\cmidrule(lr){2-4}\cmidrule(lr){5-7}
Method & $R_{\max}$ & $PT_{\max}$ & True & $R_{\max}$ & $PT_{\max}$ & True \\
\midrule
Uncompressed & 187 & 114.8 & \textbf{15.06}
             & 411 & 92.3  & \textbf{20.60} \\
\method-16k  & 132 & 28.0  & 14.94
             & 55  & 32.6  & 14.58 \\
\method-8k   & \textbf{118} & \textbf{21.0} & 14.10
             & \textbf{31}  & \textbf{20.5} & 18.55 \\
\bottomrule
\end{tabular*}
\end{table}

\subsection{Ablations}
\label{sec:experiments-ablation}

We ablate the three additions of \Cref{eq:residual} on Qwen3-8B/BCP;
full breakdown in \Cref{appendix:component-ablation}. Cross-turn
memory is the dominant signal: replacing $\mathbf{M}_t$ with the
current query reduces True~Acc by $3.13$ points at $C{=}16$k and
$1.33$ at $C{=}8$k. Cross-attention and the learned residual each
contribute under $1$ point, with the residual adding $+0.60$
True~Acc at $16$k while the rule prior is preferred by $0.60$ points
at the tighter $8$k budget---the rule scorer saturates the
dead-slot ceiling at $8$k, while the residual lifts only once
$16$k frees spare capacity. A TrimKV head-to-head and FRAMES
cross-benchmark validation appear in
\mbox{\Cref{appendix:design-axis,appendix:frames-validation}}.

\section{Conclusion}
\label{sec:conclusion}

Multi-turn agents need KV compression that survives cross-turn
intent shifts and composes with radix prefix reuse. \method
factors pruning into a cross-turn QueryMemory retention head and
a sentinel-slot layout. On Qwen3-8B/BCP this recovers Full-cache
True~Acc within $0.96$ points and holds a $20.7\%$ prefix-hit rate
where compaction baselines collapse to $0$--$3\%$; the same
dead-slot substrate transfers $44$--$46\%$ wall-time savings to
SnapKV/H2O. Retention and layout are modular axes, and the
substrate accepts any future prompt-local scorer.
Per-query budget control and $\geq\!70$B backbones remain
for future work.

\section*{Limitations}
\label{sec:limitations}

\paragraph{No autonomous budget selection.}
\method uses a single global KV budget $C$ that is fixed at
deployment time and applied uniformly to every query: the method
cannot currently choose its own compression ratio from the
trajectory. Short queries therefore pay an eviction overhead they do
not need, and the small subset of queries whose peak request tokens
grow far beyond the budget (the $92$--$115$k-token tail in
\Cref{tab:worst-case-kv}) is forced through the same compression
curve as the median query. A learned per-query controller that picks
$C$ from early-turn signals, or a schedule that expands with realized
peak usage, would let \method spend memory only on queries that need
it; we leave this to future work.

\paragraph{Supervision and scope.}
The \learn residual is trained only on strict-cleaned ToolBench
multi-turn traces with substring-matched future-action labels, so
expected gains are smaller on workloads further from agentic tool
use; the rule scorer protects against regression and \method falls
back to its heuristic on single-turn inputs. \method is validated on
Qwen3-8B and Qwen2.5-14B (Neox-RoPE GQA), and the dead-slot sentinel
currently requires fp16/bf16 KV pools. Other open backbones
(Llama-3.1~\citep{grattafiori2024llama}, Mistral-Nemo/Small, Hermes-3, GLM-4, gpt-oss, Watt-Tool)
failed prerequisite agentic behaviour on BCP under our serving harness
and so cannot isolate KV-compression effects on this workload; see
\Cref{appendix:cross-arch-gap} for the smoke study and failure-mode
taxonomy. Evaluation on $\geq$70B models is left to future work.

\bibliography{custom}

\appendix

\section{Experimental Details}
\label{appendix:hyperparams}

\paragraph{Session identification and concurrency.}
The session key is resolved in priority order: an explicit
\texttt{session\_id} field; the token tuple in a designated session
span (typically the first user task message after the chat
template); a hash of the first $P\!=\!256$ input tokens; finally, a
per-request key that never matches across requests. An LRU store of
$1{,}024$ memories per compressor caps memory; concurrent requests
update disjoint keys, so multi-tenant serving never
cross-contaminates intent. The cross-turn EMA decay
(\Cref{eq:memory-update}) is $\lambda{=}0.5$ for all reported runs.

\paragraph{Serving setup.}
All BCP runs use SGLang with radix prefix caching enabled whenever the
compression layout preserves prefix identity. Compaction baselines that
left-pack K/V rows are run with radix reuse disabled after eviction to
avoid incorrect RoPE positions. \method uses the same attention kernels
as the baselines; only the slot map differs. The evaluation harness
controls this through the KV compaction mode: main comparison baselines
use compact apply paths, while \method uses the dead-slot layout.

\paragraph{Compression settings.}
All compressors are evaluated at nominal retention budgets
$C\in\{8192,16384\}$. The budget is applied to compressible history at
each pruning event, not to the entire prompt-plus-generation request.
\method forces the resolved actionable query span to remain live and
assigns the remaining budget by score. The $W{=}32$ buffer is retained
only as a bounded query-tail fallback. StreamingLLM uses sink size $4$
plus a recent window, SnapKV uses $W{=}32$ with kernel size $7$ and
max-pool smoothing, and H2O uses ratio mode with a recent-window floor.

\paragraph{Training setup.}
The residual head is trained from strict-cleaned, ToolBench-only
multi-turn trajectories with future-action labels over a five-call
lookahead window. We use AdamW with learning rate $10^{-3}$, weight
decay $0.01$, gradient accumulation $8$, $3{,}000$ sampled examples
per epoch for Qwen3-8B and $5{,}000$ for Qwen2.5-14B (the smaller
budget matches Qwen3-8B's manifest support; \Cref{sec:method-train}),
and two epochs. Loss coefficients
are $\rho=0.05$, $\gamma=0.01$, and $\eta=0.001$; the BCE positive-rate
clip is $[0.02,0.5]$. Qwen3-8B and Qwen2.5-14B each train one
$214{,}274$-parameter pruner with the same $D{=}128$ topology.

\section{Component Ablations}
\label{appendix:component-ablation}

\Cref{tab:abl-components} ablates the three additions of
\Cref{eq:residual} on Qwen3-8B/BCP (inference-time only on the same
checkpoint; the Eff. Live KV column shifts by at most $0.5$k across
variants). Cross-turn memory is the dominant signal: replacing
$\mathbf{M}_t$ with the current query reduces True~Acc by $3.13$
points at $C{=}16$k ($14.94\!\to\!11.81$) and $1.33$ at $C{=}8$k
($14.10\!\to\!12.77$). Removing cross-attention drops True~Acc by
$0.24$ and $0.49$ points, respectively. Disabling the residual
recovers \phaseone exactly: the learned correction adds $0.60$
True~Acc at $16$k, while at $8$k the rule prior is preferred by
$0.60$ points---a tight-budget regime where an unconstrained
residual can move score mass away from the rule prior.

\begin{table}[!htbp]
\centering
\footnotesize
\caption{\textbf{Component ablations on Qwen3-8B / BCP.} Same
checkpoint, inference-time config only. ``Eff. LK'' is
\emph{Eff. Live KV} (k tokens; defined in
\Cref{sec:experiments-setup}); ablations only shift accuracy.}
\label{tab:abl-components}
\setlength{\tabcolsep}{4pt}
\begin{tabular*}{\columnwidth}{@{\extracolsep{\fill}} lrrrr}
\toprule
Configuration & Raw & Compl & True & Eff. LK \\
\midrule
\multicolumn{5}{l}{\emph{$C{=}8$k}} \\
Full \learn  & 16.67 & 84.58 & 14.10 & 10.2 \\
$-$ memory   & 14.82 & 86.14 & 12.77 & 10.1 \\
$-$ x-attn   & 15.58 & 87.35 & 13.61 & 10.3 \\
$-$ residual & 16.87 & 87.11 & \textbf{14.70} & 10.7 \\
\midrule
\multicolumn{5}{l}{\emph{$C{=}16$k}} \\
Full \learn  & 17.08 & 87.47 & \textbf{14.94} & 11.8 \\
$-$ memory   & 13.59 & 86.87 & 11.81 & 11.9 \\
$-$ x-attn   & 16.29 & 90.24 & 14.70 & 11.9 \\
$-$ residual & 15.89 & 90.24 & 14.34 & 11.9 \\
\bottomrule
\end{tabular*}
\end{table}

\section{Cross-Benchmark Validation on FRAMES}
\label{appendix:frames-validation}

The main results (\Cref{sec:experiments-accuracy,sec:experiments-efficiency})
use BrowseComp-Plus as the primary benchmark. To verify that
\method's accuracy and efficiency gains transfer beyond BCP's native
agent protocol and corpus, we replicate the agent loop on
FRAMES~\citep{krishna2024fact}, an $824$-query multi-hop question
answering suite over Wikipedia developed for the
fact-retrieval / multi-hop reasoning literature. We deliberately
\emph{adapt} FRAMES rather than wholesale-port it, so the resulting
numbers measure pruner behaviour under a BCP-style agentic
distribution while keeping the FRAMES question pool, gold answers,
and gold document URLs unchanged.

\subsection{Protocol adaptation}
\label{appendix:frames-protocol}

Out of the box FRAMES is shipped as a single-shot QA TSV with no
agent harness; the original release uses a free-form ReAct loop with
\texttt{<search>}/\texttt{<answer>} text tags. On Qwen3-8B we observe
this protocol yields a $\sim$$21$\,\% \emph{unparsable} rate (the
model emits an unclosed tag and the judge sees no answer), which
artificially caps accuracy and confounds compressor effects with
parser brittleness. Our adapted FRAMES loop differs from the native
release on four axes, each chosen to bring the per-turn KV-pressure
distribution into a BCP-compatible regime:

\paragraph{(i)~Native function calling, BCP-shaped tools.}
We replace text-tag ReAct with native OpenAI Chat Completions
function calling, registering the same two tools used in our BCP
runs verbatim: \texttt{local\_knowledge\_base\_retrieval} for
snippet search and \texttt{get\_document} for full-text drill-down.
Structured tool-call \textsc{json} arguments are robust to mid-string
truncation, dropping the unparsable rate to $\sim$$0$\,\%.

\paragraph{(ii)~BCP \texttt{QUERY\_TEMPLATE} and grader.}
We reuse BCP's \texttt{Explanation / Exact Answer / Confidence}
prompt verbatim, with inline \texttt{[docid]} citations, and grade
with the same Qwen3-32B LLM judge under BCP's
\texttt{GRADER\_TEMPLATE} (\Cref{sec:experiments-setup}). This makes
the resulting accuracy numbers commensurable across BCP and FRAMES
without rubric-level confounds.

\paragraph{(iii)~Plaintext Wikipedia corpus.}
We build a corpus of $2{,}479$ Wikipedia articles---the union of
all \texttt{wikipedia\_link\_*} URLs across the $824$ test
queries---fetched via the official
\texttt{action=query\&prop=extracts\&explaintext=1} endpoint and
capped at $3{,}000$ characters per document. The cap is chosen so
that a typical \texttt{get\_document} call returns roughly
$700$--$800$ decoded tokens---substantially shorter than BCP's
$\sim$$7$--$8$k-character tool returns, but already an order of
magnitude longer than the $\sim$$400$-character Wikipedia-intro
corpus shipped with FRAMES---which is enough to make
\texttt{get\_document} a meaningful drill-down on top of the
$800$-character snippet preview rather than a no-op. After fetching,
$99.4$\,\% of documents carry non-empty text.

\paragraph{(iv)~Stress profile.}
FRAMES questions are short enough that an unconstrained agent
answers $\sim$$30$\,\% of queries in one turn, which leaves the KV
cache too shallow for compression effects to surface. We impose a
stress profile that pushes the agent into a multi-turn regime
without altering the question or answer: $k\!=\!8$ retrieved hits
per search joined by $4$ injected distractor documents drawn from
low-BM25-score regions of the corpus, a multi-hop preamble that
requires the agent to maintain an inline evidence ledger, and
minimum-action constraints of $3$ search calls plus $1$
\texttt{get\_document} call before answering. These knobs convert
FRAMES into a multi-turn KV-pressure benchmark while keeping the
gold answers and document URLs unchanged.

What is intentionally \emph{not} aligned to BCP: the corpus size
($2{,}479$ vs.\ $\sim$$100$k docs), the corpus retrieval modality
(BM25 plaintext vs.\ FAISS dense), the per-document length, and the
question pool itself. These are dataset identities of FRAMES that
the cross-benchmark check is meant to preserve.

\subsection{Substrate integration: \method's dead-slot eviction is shared by all compressors}
\label{appendix:frames-substrate}

All compression methods evaluated below---StreamingLLM, SnapKV, H2O,
and \method---execute on the same eviction substrate introduced in
\Cref{sec:method-layout}: dropped positions are rewritten to a
reserved sentinel slot whose K drives softmax weight to zero, and
the flashinfer backend strips the sentinel before attention so
decode reads only the kept entries while token-position identity
stays intact. Concretely, the harness defaults
\texttt{KV\_COMPACTION\_MODE} to \texttt{dead\_slot} for every
score-based method, and the published \texttt{compact}-layout
RoPE-reindex apply path used in the original
StreamingLLM~\citep{xiao2024efficient},
SnapKV~\citep{li2024snapkv}, and H2O~\citep{zhang2023h2o}
implementations is only retained as a reproducibility-only opt-in
(it additionally forces \texttt{--disable-radix-cache}). This means
every method in \Cref{tab:frames-12k} shares \method's memory
profile and radix prefix-cache eligibility: the only varying
component across rows is the scorer. The same property holds for
the BCP main table; we restate it here so that the FRAMES system
metrics can be read against a fair-substrate background rather than
against each baseline's original published memory layout.

\subsection{Results at \texorpdfstring{$C{=}12$k}{C=12k}}
\label{appendix:frames-results}

\Cref{tab:frames-12k} reports True~Acc, gold-document recall, and
the system-side metrics of \Cref{sec:experiments-setup} under the
four compressors plus the no-pruning Full-cache ceiling, at
$C\!=\!12{,}288$ tokens, $N\!=\!824$ per cell. The FRAMES grader
uses the same denominator convention as BCP (non-completion counted
as wrong), so the True~Acc column is directly comparable with the
BCP main table. Full-cache is budget-agnostic and serves as the
no-pruning ceiling; PT and Eff.~LK are reported in $10^3$ tokens,
Raw and Eff.~KV reads in $10^6$ tokens, Wall in seconds, and Turns
is the mean ReAct round count. Wall-clock aggregates across the
$824$ queries with deterministic decoding ($T{=}0$), at most $32$
tool-use turns, and an agent thread parallelism of $8$.

\begin{table*}[!htbp]
\centering
\footnotesize
\caption{\textbf{FRAMES at \texorpdfstring{$C{=}12$k}{C=12k}.} Best
non-Full-cache value per column in \textbf{bold}; lower is better in
system-cost columns.}
\label{tab:frames-12k}
\setlength{\tabcolsep}{4pt}
\begin{tabular*}{\textwidth}{@{\extracolsep{\fill}} llrrrrrrrr}
\toprule
Model & Method & True Acc & Recall & PT & Eff.~LK & Raw Rd & Eff.~Rd & Wall & Turns \\
\midrule
\multirow{5}{*}{\textbf{Qwen3-8B}}
  & Full-cache                 & 28.52          & 82.13          & 17.6          & 17.8          & 15.5          & 15.6          & 68.7          & 9.23 \\
  & StreamingLLM               & 27.91          & 82.48          & 15.4          & 15.6          & \textbf{8.3}  & 8.4           & 33.6          & 8.50 \\
  & SnapKV                     & 29.00          & \textbf{82.67} & 16.8          & 17.0          & 12.9          & 13.0          & 48.6          & 8.91 \\
  & H2O                        & 27.18          & 83.00          & 16.8          & 17.0          & 12.8          & 12.9          & 46.0          & 8.94 \\
  & \method-\learn (ours)      & \textbf{29.25} & 82.11          & \textbf{12.8} & \textbf{12.9} & \textbf{6.8}  & \textbf{6.0}  & \textbf{28.9} & \textbf{7.53} \\
\midrule
\multirow{5}{*}{\textbf{Qwen2.5-14B}}
  & Full-cache                 & 34.71          & 86.60          & 23.9          & 24.3          & 39.6          & 39.9          & 236.4         & 8.86 \\
  & StreamingLLM               & 29.85          & 85.16          & \textbf{15.9} & \textbf{16.4} & \textbf{22.6} & 23.0          & \textbf{83.4} & \textbf{6.95} \\
  & SnapKV                     & 33.25          & 86.63          & 23.8          & 24.2          & 39.9          & 40.3          & 86.8          & 8.82 \\
  & H2O                        & 33.62          & \textbf{86.87} & 22.8          & 23.7          & 52.1          & 57.0          & 109.7         & 8.66 \\
  & \method-\learn (ours)      & \textbf{34.95} & 85.71          & 16.6          & 16.7          & 18.7          & \textbf{13.9} & 76.8          & 7.14 \\
\bottomrule
\end{tabular*}
\end{table*}

Two effects stand out. First, on both backbones \method-\learn
matches or exceeds Full-cache accuracy while substantially reducing
per-query KV reads: $+0.73$ True~Acc with $2.6\times$ fewer
effective KV reads on Qwen3-8B ($6.0$ vs.\ $15.6$\,M tokens), and
$+0.24$ True~Acc with $2.9\times$ fewer effective KV reads on
Qwen2.5-14B ($13.9$ vs.\ $39.9$\,M tokens). This is consistent with
the stress-profile distractor injection making the attended-to set
noisier than the unprejudiced model can profitably use, so that
retention-aware pruning acts as implicit context denoising in this
regime. Second, against the strongest heuristic baseline,
\method-\learn dominates every system axis on Qwen3-8B ($-24$\,\%
PT, $-24$\,\% effective live KV, $-54$\,\% effective KV reads,
$-41$\,\% wall-clock, $-15$\,\% ReAct turns), while on Qwen2.5-14B
it retains the headline accuracy lead and reduces effective KV
reads by $-65$\,\% relative to H2O at comparable wall-clock.
StreamingLLM's lower system numbers on the 14B row come at the
price of $-5.10$ True~Acc against the Full-cache ceiling,
illustrating the accuracy/efficiency frontier that the headline
metric obscures.

\FloatBarrier

\section{Comparison with TrimKV: Query-Decay vs.\ Positional Time-Decay}
\label{appendix:design-axis}

We compare \method-\learn against a public TrimKV~\citep{bui2025cache}
checkpoint that uses a \emph{positional time-decay} retention prior
trained for math chain-of-thought reasoning, while \method uses
\emph{query-decay} conditioned on cross-turn QueryMemory. Both are
evaluated on Qwen3-8B/BCP at $C{=}8$k with a matched Eff. Live KV
budget ($\sim$$10.1$--$10.2$k tokens) so that any accuracy
difference reflects the retention prior, not the budget.
\Cref{tab:design-axis} shows that TrimKV achieves slightly higher
per-completion correctness ($21.25$ Raw vs.\ $16.67$), but its
completion rate collapses to $51.57\%$ (vs.\ $84.58\%$ for \method).
The result is a $3.14$-point True~Acc gap
($10.96\!\to\!14.10$) once non-completion is counted as wrong,
indicating that on multi-turn tool-use trajectories—where the most
useful tokens are often the oldest (initial request, early
retrieval, first tool result)—a time-decay prior tuned for locally
relevant chain-of-thought reasoning misaligns with future
tool-call evidence, while QueryMemory-driven retention tracks it.

\begin{table}[!htbp]
\centering
\footnotesize
\caption{\textbf{TrimKV vs.\ \method at $C{=}8$k.} ``Eff. LK'' is
\emph{Eff. Live KV} (k tokens; defined in
\Cref{sec:experiments-setup}).}
\label{tab:design-axis}
\setlength{\tabcolsep}{4pt}
\begin{tabular*}{\columnwidth}{@{\extracolsep{\fill}} lrrrr}
\toprule
Configuration & Raw & Compl & True & Eff. LK \\
\midrule
TrimKV          & 21.25 & 51.57 & 10.96 & 10.1 \\
\method-\learn  & 16.67 & \textbf{84.58} & \textbf{14.10} & 10.2 \\
\bottomrule
\end{tabular*}
\end{table}

\section{Layout Ablation: Compact vs.\ Dead-Slot}
\label{appendix:layout-ablation}

\Cref{tab:abl-layout} isolates \method's hole-preserving (dead-slot)
eviction substrate from QueryMemory scoring. We re-run SnapKV
(observation-window scoring) and H2O (heavy-hitter scoring) on
Qwen3-8B / BCP at $C{=}8$k under two apply paths: the standard compact
(RoPE-reindex) path used in each method's published implementation,
and \method's dead-slot path, in which dropped positions are
redirected to a sentinel slot whose K is filled with $-10^{4}$
(\Cref{sec:method-layout}). Budget, scoring kernel, attention backend,
prefix-protect marker, and agent threading are held fixed across the
four heuristic rows; only the eviction substrate varies.

Switching the substrate cuts wall-time $44$--$46\%$ and Raw KV Reads
$39$--$47\%$ on both baselines, while Eff.\ KV Reads moves by less
than $10\%$---confirming that the wall-time savings flow through
preserved radix-prefix reuse, not tighter retention. True~Acc rises
by $2.4$--$5.5$ points on both baselines, since radix-prefix reuse
keeps per-turn ReAct prompts short and trajectories reach a final
answer before exhausting \texttt{max\_tokens}. Under the matched
dead-slot substrate, \method-\learn retains a $1.5\times$ reduction
in Raw and Eff.\ KV Reads over SnapKV ($31.6$ vs $46.9$M Raw; $39.8$
vs $59.0$M Eff.) at True~Acc within $1\sigma$ ($14.10$ vs $14.70$;
$\Delta{=}0.35\sigma$ on $830$ queries), isolating QueryMemory's
retention-side gain on top of the layout substrate.

\begin{table}[!htbp]
\centering
\footnotesize
\caption{\textbf{Dead-slot eviction transfers $44$--$46\%$
wall-time and $39$--$47\%$ Raw KV Reads savings to H2O / SnapKV
under matched scoring.} Qwen3-8B / BCP at $C{=}8$k; only the
eviction substrate varies. Wall in seconds; Raw / Eff.\ in
$10^{6}$ KV reads; True in \%.}
\label{tab:abl-layout}
\setlength{\tabcolsep}{4pt}
\begin{tabular*}{\columnwidth}{@{\extracolsep{\fill}} ll rrrr}
\toprule
Method & Layout & Wall & Raw & Eff. & True \\
\midrule
H2O    & compact   & 281.4 & 77.2 & 59.7 &  9.76 \\
H2O    & dead-slot & \textbf{150.9} & \textbf{40.7} & 54.1 & \textbf{12.17} \\
\midrule
SnapKV & compact   & 279.3 & 77.1 & 59.4 &  9.16 \\
SnapKV & dead-slot & \textbf{155.5} & \textbf{46.9} & 59.0 & \textbf{14.70} \\
\midrule
\method-\learn & dead-slot & \textbf{140.4} & \textbf{31.6} & \textbf{39.8} & 14.10 \\
\bottomrule
\end{tabular*}
\end{table}

\section{Training Curves}
\label{appendix:training-curves}

\begin{figure}[!htbp]
\centering
\includegraphics[width=\linewidth]{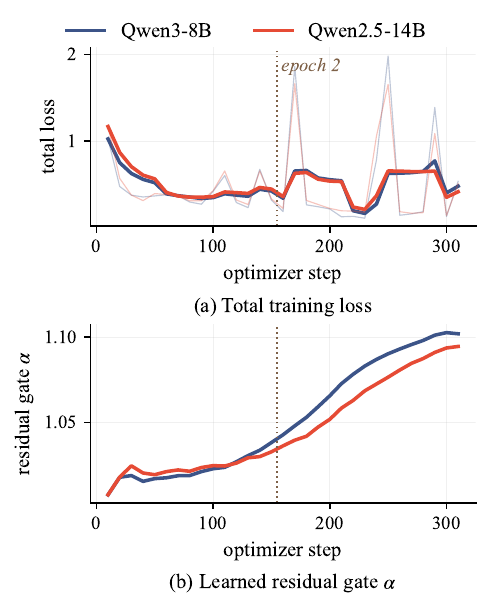}
\caption{\textbf{Training trajectories for the two trained
pruners.} Loss falls quickly and remains bounded; the learned residual
gain $\alpha$ rises from about $1.0$ to $1.1$ on both models.}
\label{fig:training-curves-xattn}
\end{figure}

\Cref{fig:training-curves-xattn} plots the optimisation
trajectories for the two \learn pruners released with \method:
the BCE loss converges within the first half of epoch~1 and
remains bounded across epoch~2, and the clipped residual gain
$\alpha$ drifts only mildly from its initialisation, consistent
with the rule-prior floor preventing the residual from dominating
the scorer.

\section{Cross-Architecture Evaluation Gap}
\label{appendix:cross-architecture}
\label{appendix:cross-arch-gap}

\subsection{Motivation}

To check whether \method generalises beyond the Qwen family used in
the main evaluation, we ran smoke tests on seven additional
open-source instruction-tuned models in the 7B--20B range, covering
four base-model families: Meta Llama (including the NousResearch
Hermes-3 fine-tune and the watt-ai watt-tool-8B tool-specialist
fine-tune, both built on Llama-3.1-8B-Instruct), Mistral, OpenAI
gpt-oss, and ZhipuAI GLM.
\textbf{Only Qwen-family models reliably engage in multi-step
autonomous tool-calling on BrowseComp-Plus} under our serving
harness; the other backbones fail at the agentic-behavior
level---not at \method{}'s KV-compression level---which prevents a
meaningful comparison of compression methods on them.

\subsection{Empirical Findings}

\Cref{tab:cross-arch-smoke} summarizes the smoke-test results.
``Avg.\ tools'' is per-query tool-call count averaged over the smoke
set; ``\textsc{Acc}'' is the judged accuracy on the full $830$-query
BrowseComp-Plus split when smoke succeeded.

\begin{table}[!htbp]
\centering
\footnotesize
\caption{\textbf{Cross-architecture smoke results on BrowseComp-Plus.}
Models marked $\times$ fail at the agentic-behavior layer and cannot
isolate KV-compression effects; failure mode is summarised in the
last column.}
\label{tab:cross-arch-smoke}
\setlength{\tabcolsep}{2pt}
\begin{tabular*}{\columnwidth}{@{\extracolsep{\fill}} lcccl}
\toprule
Model & Status & Tools & Acc & Failure \\
\midrule
Qwen3-8B-Instruct        & \checkmark & $3.3$ & $15.5\%$ & --- \\
Qwen2.5-14B-Instruct     & \checkmark & $3.3$ & ---     & --- \\
Qwen2.5-7B-Instruct      & \checkmark & $3.3$ & ---     & --- \\
\midrule
watt-tool-8B             & $\times$ & $2.0$ & $4.1\%$ & 1-shot bias \\
Llama-3.1-8B-Instruct    & $\times$ & $0.0$ & ---     & No tools \\
Mistral-Nemo-2407        & $\times$ & $0.0$ & ---     & No tools \\
Mistral-Small-3.2-24B    & $\times$ & $0.0$ & ---     & No tools \\
Hermes-3-Llama-3.1-8B    & $\times$ & $0.0$ & ---     & Forced-only \\
gpt-oss-20b (MXFP4)      & $\times$ & ---   & ---     & No FP4 HW \\
gpt-oss-20b-bf16         & $\times$ & $0.0$ & ---     & No tools \\
GLM-4-9B-Chat            & $\times$ & $0.0$ & ---     & Parser \\
\bottomrule
\end{tabular*}
\end{table}

\subsection{Failure Mode Taxonomy}

We categorize the observed failures into four structural types.

\paragraph{F1: Refusal to use tools.}
\textit{Llama-3.1-8B-Instruct} and \textit{Mistral-Nemo-Instruct-2407}
load and respond fluently, but \emph{never} emit a tool-call token on
BCP queries; they instead hallucinate direct answers from parametric
memory. As an example, the correct answer to one query is a specific
Florence hat-shop brand from the 1930s, yet Mistral-Nemo answers
``Explanation: The individual who meets all the given criteria is
\textbf{Dr.\ Jane Smith}. \ldots'' --- no retrieval was attempted; the
entity was fabricated. We attribute the gap to differences in
post-training: Qwen's instruction tuning explicitly includes multi-turn
ReAct~\citep{yao2022react} trajectories and agentic data, whereas the Llama-3.1 and
Mistral-Nemo post-training prioritizes single-turn helpfulness.

The pattern persists at larger scale and despite native tool-call
training. \textit{Mistral-Small-3.2-24B-Instruct-2506}---a
24B-parameter model that ships an
\texttt{[AVAILABLE\_TOOLS] / [TOOL\_CALLS] / [TOOL\_RESULTS]} dialogue
protocol and a Tekken tokenizer with built-in function-calling
support---also returns zero tool calls on the BCP smoke set across
three prompt configurations (default, ``MUST search'' system prompt,
and the WebSailor agent persona at temperature $0.7$). In a controlled
probe outside the BCP harness, the same model \emph{does} emit a
well-formed \texttt{[TOOL\_CALLS]} on a short single-fact query
(``find the SGLang v0.4 release date''), so the failure is not a lack
of capability but a lack of propensity to initiate retrieval when the
question is presented in BCP's long multi-criterion format. To
eliminate infrastructure as a confounder we resolved a SGLang
$\leftrightarrow$ mistral-common interop bug (the OpenAI protocol
layer generated \texttt{call\_<24hex>} tool-call IDs that
mistral-common's request validator rejects as non-conformant to its
9-character alphanumeric schema, breaking the second turn of every
multi-turn loop with a $500$). After this fix multi-turn round-trips
succeed end-to-end at the protocol level, yet the model still does not
spontaneously initiate retrieval on BCP queries---confirming that the
gap sits at the agentic-behavior layer, not at the serving layer.

\paragraph{F2: Single-shot tool use.}
\textit{watt-tool-8B}, a specialist fine-tune of
Llama-3.1-8B-Instruct for BFCL-style function
calling~\citep{patil2025berkeley}, exhibits a rigid behavior pattern:
exactly one retrieval call followed by one document fetch followed by
a final answer (avg.\ $1.0$ retrieval + $1.0$ fetch per query). The
pattern reflects the BFCL training data distribution, which is
dominated by single-turn and two-turn benchmarks, and cannot support
BCP's typical 3--5 hop information need. The resulting $4.1\%$
True~Acc—well below the $15.5\%$ Qwen3-8B reference in
\Cref{tab:cross-arch-smoke} under the same harness—demonstrates
that specialized function-calling fine-tunes can lock the model
into a single-turn behaviour profile incompatible with multi-hop
agent workloads, even when the upstream base (Llama-3.1-8B,
itself an F1 zero-tool case above) lacks autonomous tool use to
begin with.

\paragraph{F3: Hardware-bound quantization mismatch (not a model
deficiency).}
OpenAI's \texttt{gpt-oss-20b} is released in MXFP4 and requires
Hopper-class (H100/H200/B200) tensor cores for native FP4 execution;
\emph{our 80\,GiB A100 cluster has no FP4 tensor cores and therefore
cannot serve the native release at all}, so it appears in
\Cref{tab:cross-arch-smoke} as the first row \texttt{gpt-oss-20b
(MXFP4)} with the \textsc{Tools} column left blank---the model was
literally not runnable in our environment. Lacking access to
Hopper-class hardware, we additionally tried the
\texttt{lmsys/gpt-oss-20b-bf16} community MXFP4 $\rightarrow$ bf16
re-upload as a proxy (second row). The proxy never emits a tool call
on BCP queries and produces only short free-text refusals that
degenerate into repetitive token loops on continuation (e.g.,
``[\,!\,] [\,!\,] [\,!\,] \ldots''), which we attribute to precision
loss during the community dequantization---specifically in the layers
that emit OpenAI's Harmony channel-control tokens
(\texttt{<|channel|>analysis|>}, \texttt{<|channel|>final|>}). Both
rows therefore reflect infrastructure constraints on our side rather
than deficiencies in the underlying model; the native MXFP4 release
remains untested on BCP and is not evidence for or against
gpt-oss-20b's BCP capability.

\paragraph{F4: Parser / chat-template incompatibility.}
\textit{GLM-4-9B-Chat} produced parser-level garbage in our SGLang
deployment. The model emits structured tool calls in its own
proprietary format, but the \texttt{glm45} parser in SGLang is
calibrated for the GLM-4-MoE / GLM-4.5 release---a different
tokenizer and chat-template family. Adapting the parser to the
GLM-4-9B dense variant would require approximately 100--150 lines of
detector code plus chat-template adaptation, which we leave to future
work. The same parser-incompatibility risk applies to other promising
agent-tuned models (Cohere Command R-7B, IBM Granite-3.1-Instruct,
InternLM2.5) that currently lack native SGLang detectors.

\subsection{Forcing-Prompt Robustness Probe}
\label{appendix:cross-arch-forcing}

The four failure modes above are observed under each model's default
system prompt. We separately check whether stronger system prompts
recover F1-class refusals, since otherwise the cross-architecture
gap could be dismissed as a prompt-engineering artifact.
We re-ran two representative models under explicit forcing
variants---a ``you MUST call search before answering'' system
message, the WebSailor agent persona, and a
\emph{tool-choice}-style preamble that directly instructs the model
to emit a tool call on the first turn. \textit{Mistral-Small-3.2-24B}
returned zero tool calls across all forcing variants on the same
three smoke queries on which it failed by default---under temperature
$0.0$ the model collapses into degenerate template loops
(``\texttt{Your Answer: \{your answer\} | Confidence: ...}'') and
under temperature $0.7$ it still apologizes without retrieving.
\textit{Hermes-3-Llama-3.1-8B} (NousResearch), a Llama-3.1-8B
fine-tune marketed for tool use, behaved more leniently but still
unreliably: across seven prompt configurations (plain, chat-template
override, \emph{tool-choice} forcing, \emph{direct-prompt}, and
\emph{must-search}) the best result was a single tool call on
$1$ of $3$ smoke queries (\emph{direct-prompt}), with
\emph{must-search} yielding zero across all three. Neither model
approached the $3$+ spontaneous calls per query that Qwen issues at
default prompting. We therefore include the Hermes-3 row as
\emph{Forced-only} in \Cref{tab:cross-arch-smoke} to signal that any
tool-call activity observed required prompt-level coercion, and we
read this collectively as evidence that the cross-architecture gap
is not a prompt-engineering artifact: where forcing prompts work in
isolation, they do not produce the consistent multi-turn retrieval
behaviour that BCP measures.

\subsection{Why This Gap Exists}

The common pattern across F1--F4 is that BCP demands a behavior
profile---\textit{autonomous multi-step retrieval over multi-hop
trivia questions}---which few open-source post-training recipes
explicitly target. Existing tool-call benchmarks such as BFCL are
dominated by single-turn and two-turn function-selection tasks where
the system prompt already implies which tool to call; they do not
stress the agent's ability to \emph{initiate} retrieval
\emph{spontaneously} when faced with a knowledge gap. The Qwen team's
post-training, by contrast, explicitly includes ReAct-style multi-turn
traces, ToolLLaMA-derived agentic data, and BrowseComp-adjacent
research-style trajectories.

Notably, this gap is \emph{not} a deficiency in the base models'
language understanding: every failed candidate passes BFCL or similar
benchmarks with reasonable scores. The gap is in
\textit{distribution-conditional agentic behavior}, which appears to
require explicit training-data investment that, among the open-source
families we tested, only Qwen has made.

\subsection{Caveats}

We emphasize three points.

\begin{itemize}[itemsep=2pt,parsep=0pt,topsep=2pt]
\item \textbf{The claim is restricted to our sample.}
We tested seven non-Qwen models; this is not an exhaustive evaluation of
the open-source ecosystem. Closed-source models (GPT-4.1, Claude,
Gemini) likely exhibit different behavior, but their closed nature
prevents the KV-cache-level access required by our method.

\item \textbf{Prompt-engineering mitigations are orthogonal and, in
our experience, largely ineffective.}
We empirically attempted aggressive system prompts (``You MUST call
retrieval at least three times before answering''; the WebSailor
agent persona; tool-choice-style directives) on the F1 family and
found that they do not reliably recover BCP behavior in the failing
models---Mistral-Small-3.2-24B remained at zero tool calls across
all forcing variants, and Hermes-3 reached at most a single call on
$1$ of $3$ smoke queries (\Cref{appendix:cross-arch-forcing}). In
principle, even sufficiently aggressive forcing could elicit some
tool use; however, the resulting traces would no longer be comparable
across models---they would reflect the strength of the forcing prompt
rather than the model's intrinsic agentic capability, conflating
prompt-engineering effects with \method{}'s intrinsic benefit. We
therefore exclude both natural and forced runs of the failing models
from the main evaluation.

\item \textbf{This is not a critique of the failing models.}
Llama-3.1, Mistral-Nemo, GLM-4-9B-Chat, and others remain capable
general-purpose assistants; their training simply does not target the
specific agentic behaviour BCP requires. Closing this gap in future
open-source releases would allow a more thorough test of
\method{}'s backbone-agnostic claim.
\end{itemize}

\section{Extended Algorithmic Details}
\label{appendix:algos}

\paragraph{Compression event.}
Given request $r$, slot map $\mathbf{S}_r[0{:}N)$, budget $C$ and
dead slot $s^\dagger$, \method skips compression when $N\le C$.
Otherwise it forms the live-position set
$\mathcal{L}=\{j:\mathbf{S}_r[j]\ne s^\dagger\}$, resolves the
decision query span, and splits $\mathcal{L}$ into forced positions
$\mathcal{F}$ and candidate positions $\mathcal{A}$.
For \texttt{query}, the scorer uses only the current
$\mathrm{Enc}(\mathbf{q}_t)$.
For \texttt{memory} and \texttt{learnable}, the scorer first derives a
session id, updates $\mathbf{M}_t$, computes
\Cref{eq:rule-score}, and optionally adds the residual
\Cref{eq:residual}.
Tensor-parallel ranks all-reduce the resulting score vector.
The kept set is
$\mathcal{K}=\mathcal{F}\cup\mathrm{topk}(\mathcal{A},
\max(0,C-|\mathcal{F}|))$.
The implementation persists a boolean valid mask, redirects newly
dropped positions to $s^\dagger$, and leaves all kept slot ids
unchanged.

\paragraph{Prefix-safe free.}
For each newly dropped position $j$, the old slot
$u=\mathbf{S}_r[j]$ is returned to the allocator only if
$j$ is outside the radix-protected prefix, $u\ne s^\dagger$, and
$u$ is absent from the final kept-slot set
$\{\mathbf{S}_r[i]:i\in\mathcal{K}\}$.
This alias check is required because a request may already contain
redirected positions or shared prefix slots.

\paragraph{Session id derivation.}
The session key is chosen by the first available source:
an explicit \texttt{session\_id}, then the token tuple in a designated
session span, then a hash of the first $P$ input tokens.
If the request has no token ids, the implementation uses a
request-local key; this preserves correctness by avoiding accidental
memory sharing across unrelated embedding-input requests.

\paragraph{\learn training step.}
Per training row, the trainer extracts post-RoPE K and query-span
Q from the frozen model, averages the Q rows to form
$\mathrm{Enc}(\mathbf{q}_t)$, and updates the same
\texttt{QueryMemory} used at inference. It then computes the rule
score and residual features over the remaining candidate
positions, builds the cross-attention memory from query-span
K-vectors, obtains MLP logits, and applies \Cref{eq:loss}
(session replay order matches \Cref{sec:method-train}).

\section{Artifact Licenses and Intended Use}
\label{appendix:artifact-licenses}

We summarise the licenses, intended use, and our use of every
external artifact referenced in the main text and appendices.
All artifacts are used in accordance with their original licenses
and for research purposes consistent with their stated intended use.

\paragraph{Language models.}
\textbf{Qwen3-8B}~\citep{yang2025qwen3} and
\textbf{Qwen2.5-14B}~\citep{qwen2025qwen25technicalreport}, used as
the primary backbones, are released by Alibaba under the Apache
License 2.0; \textbf{Qwen3-32B}, used as the LLM-as-a-judge grader,
is released under the same license. The smoke-test models in
\Cref{appendix:cross-architecture} carry the following licenses:
\textbf{Llama-3.1-8B-Instruct}~\citep{grattafiori2024llama} under
the Llama 3.1 Community License Agreement;
\textbf{Hermes-3-Llama-3.1-8B} (NousResearch) inherits the same
Llama 3.1 Community License;
\textbf{Mistral-Nemo-Instruct-2407} and
\textbf{Mistral-Small-3.2-24B-Instruct-2506} under the Apache
License 2.0; \textbf{watt-tool-8B} under the Apache License 2.0
(its model card declares Apache-2.0; the upstream base model is
Llama-3.1-8B-Instruct, which itself carries the Llama 3.1 Community
License); \textbf{gpt-oss-20b} under the Apache License 2.0;
\textbf{GLM-4-9B-Chat} under \textit{The glm-4-9b License} (HF
license tag \texttt{glm-4}).
All models are used for inference-only evaluation; the base
weights are never updated. Our use is consistent with each
model's release statement of academic/research use.

\paragraph{Datasets and benchmarks.}
\textbf{BrowseComp-Plus}~\citep{chen2025browsecomp}, the primary
evaluation suite, is released under the MIT License; we use the
public 830-query split and the released document corpus without
modification beyond the per-query retrieval cache described in
\Cref{sec:experiments-setup}.
\textbf{FRAMES}~\citep{krishna2024fact} (Google) is released under
the Apache License 2.0; we use its 824-query test split, gold
answers, and \texttt{wikipedia\_link} URLs unchanged
(\Cref{appendix:frames-protocol}).
\textbf{ToolBench}~\citep{qin2024toolllm} and its strict-cleaned
re-release \textbf{StableToolBench}~\citep{guo2024stabletoolbench}
are released under the Apache License 2.0 and used only for
training the residual head (\Cref{sec:method-train}); no ToolBench
content appears in the evaluation distribution.
The Wikipedia corpus constructed in \Cref{appendix:frames-protocol}
is retrieved from the official MediaWiki API and is governed by the
\textbf{CC BY-SA 4.0} license; we redistribute only the URL list
and SHA-256 hashes of the cached extracts rather than the article
text itself.

\paragraph{Software and serving stacks.}
\textbf{SGLang}~\citep{zheng2024sglang} and
\textbf{vLLM}~\citep{kwon2023efficient}, on which our compression
substrate is implemented and evaluated, are released under the
Apache License 2.0. The reference \textbf{TRIM-KV} checkpoint
\texttt{ngocbh/TrimKV-Qwen3-8B-Math}~\citep{bui2025cache} used in
\Cref{appendix:design-axis} is distributed by its authors on
Hugging Face under the Apache License 2.0.

\paragraph{Released artifact.}
Our code and trained residual-head checkpoints will be released
under a permissive open-source license (MIT) consistent with our
upstream dependencies. The release contains only model weights for
the residual head (a $214{,}274$-parameter MLP) and integration
code, with no redistribution of the base LLM weights, ToolBench
corpus, or BrowseComp-Plus content.

\section{AI Assistant Use Statement}
\label{appendix:ai-assistant-use}
AI assistants were used for language polishing, checklist
interpretation, and minor writing assistance. All technical claims,
experimental results, code, and final submission content were
reviewed and approved by the authors.

\end{document}